\pdfoutput=1

\documentclass[11pt]{article}

\usepackage[]{acl}

\usepackage{times}
\usepackage{latexsym}

\usepackage[algoruled, boxed, vlined, linesnumbered]{algorithm2e}

\usepackage[T1]{fontenc}

\usepackage[utf8]{inputenc}

\usepackage{microtype}

\usepackage{bm}
\usepackage{multirow}
\usepackage{subfig}
\usepackage{algorithmic}
\usepackage{amsfonts}
\usepackage{booktabs}
\usepackage{graphicx}
\usepackage{comment}
\usepackage{hyperref}

\title{Controllable Fake Document Infilling for Cyber Deception}

\author{Yibo Hu, Yu Lin, Erick Skorupa Parolin, Latifur Khan, Kevin Hamlen\\ 
    The University of Texas at Dallas\\
    \normalsize{\texttt{\{yibo.hu,yxl163430,erick.skorupaparolin,lkhan,hamlen\}@utdallas.edu}}\\
}

\begin{document}
\maketitle   
\begin{abstract}
Recent works in cyber deception study how to deter malicious intrusion by generating multiple fake versions of a critical document to impose costs on adversaries who need to identify the correct information.
However, existing approaches are context-agnostic,  resulting in sub-optimal and unvaried outputs.
We propose a novel context-aware model, Fake Document Infilling (FDI), by converting the problem to a controllable mask-then-infill procedure.
FDI masks important concepts of varied lengths in the document, then infills a realistic but fake alternative considering both the 
previous and future contexts. 
We conduct comprehensive evaluations on technical documents and news stories. 
Results show that FDI outperforms the baselines in generating highly believable fakes with moderate modification to protect critical information and deceive adversaries.
\end{abstract}

\section{Introduction} 
\label{sec:INTRODUCTION}
\begin{table*}[t]
\centering
\small
\scalebox{0.9}{
\begin{tabular}{p{15.5cm}}
\toprule
\textbf{A. Original Article } 
Tomographic Image Reconstruction using Training images 

We describe and examine an algorithm for tomographic image reconstruction where prior knowledge about the solution is available in the form of training images. 
We first construct a nonnegative dictionary based on prototype elements from the training images; this problem is formulated as a regularized non-negative matrix. Incorporating the dictionary as a prior in a convex reconstruction problem, we then find an approximate solution with a sparse representation in the dictionary...
\\
\midrule
\textbf{B. GPT-2: Generation given a prompt} 
...We describe and examine an algorithm for tomographic image reconstruction where prior knowledge about the solution is available in the form of training images. 
\textcolor{red}{Instances were reconstructed from their images using image and pixel centroids. The concept of image reconstruction provides several advantages over previous techniques, such as indexing the solution to a representation with integral or submaximal number of cepstrates,} ...
\\
\midrule
\textbf{C. WEF-Replacing nouns} 
...We first construct a nonnegative \textcolor{blue}{dictionary }based on prototype elements from the \textcolor{blue}{training }images; this problem is formulated as a regularized non-negative matrix \textcolor{blue}{factorization}. Incorporating the \textcolor{blue}{dictionary }as a prior in a convex reconstruction problem, we then find an approximate \textcolor{blue}{solution }with a sparse representation in the \textcolor{blue}{dictionary}...
\\
\textbf{D. WEF-Generation} 
...We first construct a nonnegative \textcolor{red}{sparsity }based on prototype elements from the \textcolor{red}{encoderdecoder }images; this problem is formulated as a regularized non-negative matrix \textcolor{red}{orthonormal}. Incorporating the \textcolor{red}{sparsity }as a prior in a convex reconstruction problem, we then find an approximate \textcolor{red}{strategy} with a sparse representation in the \textcolor{red}{sparsity}...
\\
\midrule
\textbf{E. FDI-Replacing $n$-grams} 
...We first construct \textcolor{blue}{a nonnegative dictionary }based on \textcolor{blue}{prototype elements} from the training images; this problem is formulated as a regularized non-negative matrix factorization. Incorporating the dictionary as a prior in \textcolor{blue}{a convex reconstruction problem}, we then find \textcolor{blue}{an approximate solution with a sparse representation }in the dictionary...
\\
\textbf{F. FDI-Generation} 
...We first construct \textcolor{red}{a collection of missing patches} based on \textcolor{red}{images} from the training images; this problem is formulated as a regularized non-negative matrix factorization. Incorporating the dictionary as a prior in \textcolor{red}{the whole dictionary}, we then find \textcolor{red}{a similar estimate for missing patches} in the dictionary...
\\
\bottomrule
\end{tabular}
}
\caption{Comparison of strategies and generated samples from different models.
We preserve the document's head (the headline and the first sentence) and modify only the document's body.
GPT-2 generates a new \textcolor{red}{body} given the document's head as a prompt. 
WE-FORGE (WEF) and FDI substitute certain \textcolor{blue}{concepts} with \textcolor{red}{ alternatives}. }
\label{tab:fake_sample_intro}
\end{table*}

According to the statement of the U.S. Securities and Exchange Commission, the scope and severity of cyber risks have dramatically increased, and constant vigilance is needed to protect against intrusion \citep{cyber}. 
Cyber Deception is a cybersecurity defense practice \cite{masud2007hybrid,tu2008secure,akbar2022knowledge} that aims at protecting critical documents once intruders penetrate the network system \citep{yuill2004honeyfiles,bowen2009baiting}.
The goal is to deceive attackers by deploying decoys such as fake documents and thus increase their cost to identify critical information.

In this work, we aim at designing a novel fake document generator that combines Cyber Deception and Natural Language Generation (NLG) technologies to generate controllable, diverse, and believable fakes at scale to protect critical information and deceive adversaries. 
Although recent works in Cyber Deception develop strategies to generate complicated fake technical documents such as patents, few consider adopting pretrained contextual features to enhance scalability and generation quality. 
For example, 
FORGE \citep{forge} generates fake documents by replacing the concepts of a technical document with semantically similar alternatives from an expensive prerequisite ontology.
WE-FORGE \citep{weforge} eliminates the need for ontologies by using word embedding distances. However, it identifies potential replacements only for unigrams (especially nouns) based on unbalanced word embedding clusters in a context-agnostic manner, resulting in sub-optimal or inadequate alternatives.

Meanwhile, recent studies in NLG have been driven by pre-trained contextual language models (LMs), which can generate increasingly realistic but less-controllable text \citep{gpt2,t5,bart,xlnet}. 
Sub-fields such as controllable text generation \citep{ctrl,pplm}, story generation \citep{clark2018neural,fan2018hierarchical}, and text infilling \citep{fedus2018maskgan,ilm} further study how to leverage LMs to generate content with desired attributes.
However, few methods offer fine-grained control over concept levels or provide an efficient, controllable fake text generation strategy.

We propose a novel context-aware model, Fake Document Infilling (FDI), by converting fake document generation into a controllable mask-then-infill procedure. 
Specifically, we select and mask essential concepts of varied lengths in a document. 
Then we infill the masked spans with realistic but fake alternatives based on contextualized knowledge from an LM. 
To the best of our knowledge, we are the first to propose a complete controllable mask-then-infill model and design a comprehensive evaluation scheme to study fake text generation.

To briefly demonstrate the motivation for this work, 
 \autoref{tab:fake_sample_intro} illustrates the difference between an LM (i.e., GPT-2 finetuned on the target dataset \citep{gpt2}), WE-FORGE (WEF), and our FDI in generating fake samples of the same document.
We preserve the document’s head (the headline and the first sentence shown in A) and modify the document’s body. 
GPT-2 generates a new \textcolor{red}{body} (shown in red in B) based on the original head in a left-to-right manner.
The output is fluent but less controllable and may gradually go off-topic.
Besides, GPT-2 cannot control text length and wrapping, hindering its application when following a layout is strictly required.

Both WE-FORGE and FDI adopt the strategy of replacing specific \textcolor{blue}{concepts} (shown in blue in C and E) in the original document with \textcolor{red}{alternatives} (shown in red in D and F). 
However, WE-FORGE suffers from three major limitations.
First, WE-FORGE needs to train word-embeddings from scratch for every custom dataset, requiring large training corpora~\citep{pennington2014glove,mikolov2018advances}.
Second, its word-embedding-based clustering of concepts is unbalanced and sensitive to the initialization and hyper-parameters, resulting in limited replacements for some given concepts.
Finally, WE-FORGE only replaces nouns and is agnostic to context, limiting the diversity and quality of the generated text.

In contrast, FDI provides many advantages over previous methods.
First, instead of training word embeddings from scratch, FDI finetunes a pre-trained LM to generate human-like text with limited data.  Furthermore, FDI replaces spans of arbitrary lengths, considering the document context (through the LM) to improve the outputs' diversity and coherency.
Finally, FDI implements strategies to select (mask) and find alternative concepts (infill), protecting essential details from original documents and producing realistic fake samples.

To validate the outperformance of FDI, we design an innovative set of experiments combining evaluation methods observed in distinct areas, i.e., cyber security and NLG.
We collect reviews from more than 40 volunteers over $1.4k$ fake documents on technical and non-technical datasets.
Finally, we compile the reviews to evaluate the model's ability to generate natural text and its effectiveness in protecting the original information and deterring attackers.
Our code is publicly available.\footnote{
\url{https://github.com/snowood1/FDI}}

\section{Related Work} 
\label{sec:RELATED WORK}
\paragraph{Cyber Deception.}
Cyber Deception aims at deceiving attackers by misguiding them toward inaccurate information with deployed decoys in the network systems of enterprises.
Early works generate decoy honey files \citep{yuill2004honeyfiles, white2006using,bowen2009baiting,whitham2013automating}
or simple documents with basic NLP methods \citep{voris2012lost,Wang2013generation} to entice attackers and improve intrusion and exfiltration detection.
Recent works combine advanced NLP techniques to generate fake technical documents at scale while enhancing believability.
These efforts include substituting words or concepts based on part-of-speech tagging~\citep{whitham2017automating}, prerequisite ontologies~\citep{forge}, concept occurrences graphs~\citep{fake_karuna2021}, or word embeddings~\citep{weforge}.
Nevertheless, these methods are context-agnostic, limiting producing diverse and natural outputs.
The only exception \citep{fake_gpt} uses vanilla contextualized LMs on short description texts instead of long technical documents.

\paragraph{Controllable Text Generation.}
Building costly conditional LMs for desired attributes, by either training from scratch \citep{Grover,ctrl} or back-propagating gradients \citep{pplm}, are extensively studied. 
The attributes are usually pre-defined by a list of control codes or keywords. 
Other lightweight alternatives are proposed by using discriminators or Bayes’ rules to control the attributes of generated text during the decoding time \citep{KrauseGeDi2020,yang2021fudge,liu2021dexperts}.  
A subfield called \textbf{Story Generation} focuses on generating short stories given hints such as title, storyline, premise, entities, or rare words \citep{clark2018neural,fan2018hierarchical, fan2019strategies, peng2019plan,goldfarb2020content,rashkin2020plotmachines,Tan2021ProgressiveGO,ippolito2019unsupervised, Ippolito2020TowardBS,das2020can}.
Specifically, \citet{Grover} generate fake news stories conditioned on metadata from a list of propaganda websites. 
Nevertheless, these fields differ from our task. They mainly focus on non-technical domains (e.g., news and stories) and lack fine-grained control over concept levels.

\paragraph{Text Infilling.}
Text infilling is a generalization of the cloze task~\citep{taylor1953cloze} from single words to spans of varied lengths.
Current works focus on correctly infilling the incomplete text for applications in text editing or ancient documents restoration \citep{fedus2018maskgan,Zhu2019TextI,liu-etal-2019-tigs,zaidi2019decoding,ilm,shen2020blank}.
However, the \textit{controllable mask-then-infill} task addressed in this paper is more complex. It involves masking relevant concepts (text spans) in a document and infilling realistic yet misleading spans to replace such masks.

\paragraph{Adversarial augmentation.}

This task aims at generating perturbed augmented samples to improve the robustness of NLP models by heuristic rules that replace words from WordNet or word embeddings 
\citep{alzantot2018generating,jia2019certified,ren2019generating,eda},
contextualized perturbations~\citep{garg2020bae,li2020bert,clare}, or comprehensive frameworks~\citep{Ribeiro2020BeyondAB,Morris2020TextAttackAF,wu2021polyjuice}.
Again, these random perturbation methods lack precise control over concepts, hindering their usage for our task.

\section{Approach} 
\label{sec:APPROACH}
\begin{figure*}[t]
  \centering
  \includegraphics[width=0.92\linewidth]{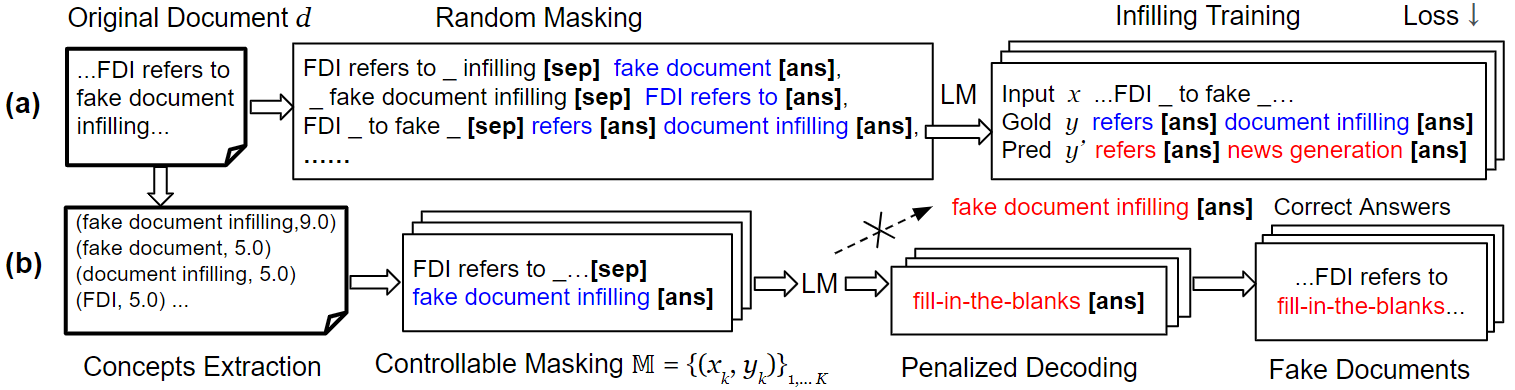}
  \caption{The (a) training and (b) inference steps of FDI. Random masking is used to train an LM for general text infilling. During inference, we mask important text spans and utilize the trained LM to replace such masked spans. Penalization mechanism is applied to encourage generating fake answers.} 
  \label{fig:Pipeline}
\end{figure*}

\subsection{Framework} 

We follow the same convention of fake document generation proposed in FORGE \citep{forge}:  
Given a real document $d$ as input, the model generates a set $D'$ of fake documents. 
Each fake document $d' \in D'$ is similar to $d$ to be believable, yet sufficiently different from $d$ to be inaccurate.
We obtain $d'$ by replacing certain concepts $c$ of $d$ with alternatives $c'$. 
High-quality $d'$ is expected to cost the attacker much time to identify the real $d$ from the $|D'|+ 1$ documents. 
Thus, a fake document generator needs to ensure believability by considering at least two aspects: (1) how to select the set of concepts $C$ to be replaced;  and (2) how to choose replacement concept $c'$ for every $c \in C$.

We convert the formulation above into the controllable document infilling task. Given a document $d$, we first extract and mask text snippets of varied lengths expressing each important concept $c \in C$. Then, we use an LM to infill the masked spans with realistic but inauthentic alternatives $c'$, considering the context of these spans.
FDI addresses these sub-tasks by designing (1) a controllable masking function to select concepts and (2) a decoding strategy to replace the masked spans.

\autoref{fig:Pipeline} shows the (a) training and (b) inference procedures of FDI.
First, we apply the random masking approach to train a robust and flexible LM to fill various types of masks. 
Then, we use a curated strategy to precisely steer text generation during the inference step. 
Specifically for inference, we first use controllable masking to produce masked examples, protecting essential information of $d$. 
Then, we use the trained LM to replace each masked concept $c \in C$ with a sampled $c'$. 
To ensure the fakeness of the generated document, we introduce a penalization factor in the decoding step to avoid the model predicting the original concept (i.e., let $c'\neq c$). 
Finally, we obtain a completed fake document by infilling the input text with the predicted alternatives. We detail each component of FDI in the following subsections.

\subsection{Training}
\label{sec:training}

The training step involves finetuning an LM to text infilling task, utilizing the random masking approach.
\autoref{fig:Pipeline}(a) illustrates three different training pairs. Each concatenates input $x$ and target $y$ by a separator token $[sep]$.
$x$ is generated by a random masking function $f(d)$, which replaces specific spans $C=\{c_1,...,c_n\}$ in document $d$ with special (blank) tokens.
$y=c_1 [ans] ... c_n [ans]$ refers to the answers to the blanks concatenated with special tokens $[ans]$.
We finetune LM($\theta$) to learn the distribution $p_{\theta}(y|x)$ by minimizing the cross-entropy loss between the target $y$ and the probability distributions of prediction $y'$. 

We design $f(d)$ to generate various masked examples with coarse control over the granularities.
Similar to \citep{ilm}, we use more special tokens instead of a universal blank to specify three granularities: words, $n$-grams, and sentences. 
For example,  $x$ in \autoref{fig:Pipeline}(a) becomes ``FDI $[masked\_word]$ to fake $[masked\_ngram]$.'' (we only show a universal blank ``\_'' in the figure for simplicity). 
Next, we traverse the hierarchy of $d$ to sample each mask type randomly and obtain a masked token rate of  15\% suggested in \citep{bert,ilm} (details in \autoref{sec:hyperparameters}).
Finally, we generate various masked examples of each $d$ for training data augmentation.

\subsection{Inference} 

The inference process (shown in \autoref{fig:Pipeline}(b)) produces fake documents through the following steps: (1) extracting and selecting the appropriate set of concepts $C$; and (2) determining the fake replacement $c'$ for every $c \in C$, through decoding method.

\subsubsection{Concepts Extraction and Selection}\label{sec:concepts extraction}

Concepts extraction and selection are essential components in fake document generators and vary in schemas. Therefore, we define the following settings:
First, instead of expensive annotation \cite{forge}, we followed recent works to use automatical keywords (e.g., based on TF-IDF \cite{weforge}) as critical information for scalable evaluation. We chose RAKE \cite{rake} to score $n$-grams to extract concepts with varying lengths without additional cost. 
Second, we only revised the document body with the head unchanged, as illustrated in \autoref{tab:fake_sample_intro}. Completely altering the head may let intruders skip the forged document quickly. We expect the fake samples to alter critical details without changing the topics. 
Besides, this setting enables us to compare naive GPT-2 that needs to initiate with a given prompt and is commonly used in NLG evaluation \cite{clark2021all}.

Algorithm \ref{alg:alg} illustrates the \textit{Controllable Masking} procedure for selecting and extracting concepts, which includes two main parts: (1) Lines \ref{alg:parse}-\ref{alg:getsent} build the candidate pool of concepts from document $d$, consisting of sets $C$ (words or $n$-grams) and $S$ (whole sentences); and (2) Lines \ref{alg:declareM}-\ref{alg:returnM} generate $K$ masked examples $\mathbb{M} =\{M_1,..., M_K\}$ by sampling from the candidate pool. 
Each masked example $M_k$ results in various fake documents during the later decoding step. 
In this way, FDI can produce diverse examples which vary in masked locations and replacements of each mask.

In the first part of Algorithm \ref{alg:alg}, Line \ref{alg:parse} splits document $d$ by stop-words and delimiters to create the initial set of concepts $C$. Line \ref{alg:rakescore} computes the importance score of each concept $c \in C$ through the degree $deg(w)$ and frequency $freq(w)$ in its word co-occurrence graph of a term $w$ occurring in $c$, following \citep{rake}.
Next, we filter the concepts based on the quantile $Q_R(\cdot)$ of the importance scores in Line \ref{alg:filterMinScore}. 
Long concepts often get higher RAKE scores than short concepts.
For example, in \autoref{fig:Pipeline} (b), ``fake document infilling'' gets a higher score than its member phrases.
Therefore, we empirically set the lower bound $q_{min}$ to $40\%$, a trade-off between concepts' importance and diversity (in terms of length).

A document’s head  (e.g., the title and the first sentence) often contains topic words and summarizes the content. Thus, we ignore extracting these phrases to prevent generating entirely off-topic articles in Line \ref{alg:filter}, as discussed in the start of \autoref{sec:concepts extraction}. 
For instance, we remove the selected topic concept ``tomographic image reconstruction'' from the candidate set in \autoref{tab:fake_sample_intro} A.

RAKE removes masked phrases' determiners and may result in obvious plural noun errors during infilling. 
For example,  LMs infill ``find an \underline{approximate solution}...''  to ``find an \underline{similar estimate}...'' in \autoref{tab:fake_sample_intro} (E).
The easiest solution to alleviate such errors is to replace the extracted span with its determiner as a whole, e.g., ``find an \underline{approximate solution}...''.
Thus, we concatenate extracted spans with their determiners through function $concatDet(\cdot)$ in Line \ref{alg:concatDet}.

\begin{algorithm}[t]
\DontPrintSemicolon
\small

\SetKwData{conceptsSet}{$C$}
\SetKwData{sentencesSet}{$S$}
\SetKwData{minimumScore}{$minScore$}
\SetKwData{minimumQuantile}{$q_{min}$}
\SetKwData{stopWords}{$W_{st}$}
\SetKwData{sentenceRatio}{$t_s$}
\SetKwData{probabilitySentences}{$p_s$}
\SetKwData{probabilityConcepts}{$p_c$}

\SetKwFunction{parse}{$splitConcepts$}
\SetKwFunction{quantileFunction}{$Q_R$}
\SetKwFunction{getSentences}{$getSents$}
\SetKwFunction{concatenateDeterminant}{$concatDet$}
\SetKwFunction{range}{$range$}
\SetKwFunction{randomSample}{$randomSample$}
\SetKwFunction{maskedRate}{$maskedRate$}
\SetKwFunction{maskFunction}{$getMaskedInput$}

\SetKwInOut{Input}{Input}
\SetKwInOut{Output}{Output}

\SetKwRepeat{Do}{do}{while}

\Input{document $d$, stop-words \stopWords, thresholds \minimumQuantile, \sentenceRatio, $\gamma$, masking probabilities \probabilitySentences, \probabilityConcepts.
}

\Output{list of masked examples $\mathbb{M}$ of size $K$.}

\conceptsSet $\gets$ \parse{d, \stopWords} \label{alg:parse}

$R_C \gets \{r(c):\sum_{w \in c} deg(w)/freq(w)$ for $c$ in \conceptsSet\}\label{alg:rakescore}

\conceptsSet $\gets\{c$ if $r(c)$ $\geq$ \quantileFunction{\conceptsSet, \minimumQuantile} for $c$ in \conceptsSet\} \label{alg:filterMinScore}

\conceptsSet $\gets\{c$ if $c \notin d.head\}$ \label{alg:filter}

\conceptsSet $\gets$ \concatenateDeterminant(\conceptsSet, $d$) \label{alg:concatDet}

\sentencesSet $\gets$ \getSentences{d, \conceptsSet, \sentenceRatio} \label{alg:getsent}

$\mathbb{M}\gets[\emptyset]$ \label{alg:declareM}

\For{$k$ in \range{$K$}}{   \label{alg:forEachK}
    $y_k \gets \{\emptyset\}$
    
    \Do{\maskedRate{d, $y_k$} $<\gamma$}{
        $y_k \gets y_k \cup \{\randomSample{\sentencesSet, \probabilitySentences} \}$
        
        $y_k \gets y_k \cup \{\randomSample{\conceptsSet, \probabilityConcepts} \}$
    }  \label{alg:sampling}
    
    $y_k \gets mergeCloseMasks \: (y_k)$ \label{alg:merge_close_concepts}
    
    $x_k \gets$ \maskFunction{d, $y_k$} \label{alg:get_x}

    $\mathbb{M} \gets \mathbb{M}$ $\cup$ ($x_k$, $y_k$) \label{alg:updateM}
    
}

\Return $\mathbb{M}$ \label{alg:returnM}

\caption{Controllable Masking}
\label{alg:alg}
\end{algorithm}

Besides candidate concepts $C$, we can optionally replace sentences with a high density of key concepts. In practice, replacing a whole sentence generally produces better results than densely infilling many blanks in the same sentence. Therefore, in Line \ref{alg:getsent}, we collect the sentences from $d$ whose percentage of tokens belonging to any concept in $C$ is higher than the threshold $t_s$. Function $getSents(\cdot)$ returns such dense sentences to form the set $S$.

Once we obtain the candidate pool of concepts from $d$, we sample $K$ masked examples $\mathbb{M} =\{M_1,..., M_K\}$. Each ${M}_k$ includes input $x_k$ with special masked tokens, and the answer spans $y_k$ (as shown in \autoref{sec:training}). In Lines \ref{alg:forEachK}-\ref{alg:sampling}, for each masked example, we collect $y_k$ by sampling the sets $S$ and $C$ with probabilities $p_s$ and $p_c$, respectively. We iteratively sample until we get enough non-overlapping concepts that reach a threshold of masked token rate $\gamma$.

We also merge short masked spans located closely within the same sentence into longer spans to reduce the number of masked spans in ${M}$ in Line \ref{alg:merge_close_concepts}. For example,  we merge the two masked spans in ``find \underline{an approximate solution} with \underline{a sparse representation} ...'' to one span in \autoref{tab:fake_sample_intro} (E).
We get the corresponding masked input $x_k$ by replacing spans in $y_k$ with special masked tokens in Line \ref{alg:get_x}. We repeat the above sampling process to get our collections of $(x_k,y_k)$ pairs.

\subsubsection{Decoding}\label{sec:decoding}

We design a penalized decoding strategy based on Top-$p$\% sampling \citep{holtzman2019curious} to generate natural yet fake texts.
The training step minimizes the cross-entropy loss between the answers and prediction probabilities to retrieve the original document.
However, the inference uses sampling instead of greedy search to get various outputs that are unlikely to be identical to the original document.
Furthermore, we discount the scores of the tokens for the correct answers to encourage fake outputs during the inference, similar to the mechanism for discouraging repetition \citep{ctrl}.

Specifically, we first get a subset of tokens $A$ from the correct answers $y$ of each $M$ by filtering out too-short tokens, probably  stopwords or insignificant sub-words such as prefixes.  
Then, given the input $\textbf{x}$, the probability distribution over the next possible token being word $i$ in the vocabulary $V$ is the softmax:
\begin{eqnarray} \label{eq:decoding} 
p(y=i|\textbf{x})=\frac{\exp \left(z_{i} /(T \cdot I(i \in A))\right.}{\sum_{j} \exp \left(z_{j} /(T \cdot I(j \in A))\right.},
\end{eqnarray} 
where $T$ is the temperature parameter and $z_i$ is each $i$'s score. $I(\cdot)=\delta$ if true else 1, and $\delta$ is the penalty parameter.
A high $\delta$ discourages generating correct answers but also produces errors.
Thus, we set $\delta=1.2$ in our experiments based on our empirical observation.
Finally, following \citep{holtzman2019curious}, we sample from the most probable tokens whose cumulative probability comprises the top-$95\%$ of the entire vocabulary.

One concern of the inference step is to control the fakeness of the output. 
Substituting concepts with similar semantic replacements fails to protect critical information.
Due to its unbalanced and unvaried candidate pool, WE-FORGE often suffers from replacing a noun concept with its synonyms, such as substituting  “solution” with “strategy” in \autoref{tab:fake_sample_intro} C and D.

In contrast, FDI controls fakeness efficiently by masking various spans from words to sentences, significantly improving the diversity and thus reducing the chance of getting similar outputs. Moreover, FDI infills fake samples conditioned on the incomplete context hiding critical information. Even for the exact phrases that occur in different places, we do not replace them with an identical replacement. Instead, the LM decodes their plausible replacements based on different contexts. This infilling and sampling process favors common, safe, but lossy answers. For example,  the document with masked concepts in \autoref{tab:fake_sample_intro} E can result in various outputs with the same structure but distinct details. Later, the sampled answers like ``images'' and ``the whole dictionary'' in \autoref{tab:fake_sample_intro} F seem natural but uninformative - they hide the critical information expressed in the original document. Finally, the penalty mechanism in \autoref{eq:decoding} also encourages the model to infill a fake answer.

\section{Experiments} 
\label{sec:EXPERIMENTS}
\subsection{Datasets}

Following previous cyber deception works, we conducted experiments on two technical datasets: the \textbf{CS} \citep{ilm} and the patent abstracts dataset (\textbf{PAT})\footnote{\url{https://github.com/chirag-choudhary/Patent-Summarizer}}. 
The first consists of abstracts from computer science papers on arXiv. The latter covers topics such as Electrical, Chemistry, and Biology. Additionally, we experimented on a non-technical dataset by crawling and filtering a subset of news from the Wall Street Journal (\textbf{WSJ}). 
\autoref{tab:datasets} summarizes three datasets' statistics, document lengths, and training sequence lengths we chose.

\begin{table}
\small
\centering
\setlength{\tabcolsep}{4pt}
\begin{tabular}{ccccc} 
\toprule
Dataset& Train / dev / test    & \# tokens & seq-len \\
\midrule
CS  & 409,555 / 8,547 / 8,498 & $205 \pm 70$ & 400 \\
PAT & 16,000 / 4,000 / 5,743  & $132 \pm 58$ & 256 \\
WSJ & 40,862 /2,270 / 2,270   & $292 \pm 78$ & 512 \\
\bottomrule
\end{tabular}
\caption{The datasets used in our experiments.}\label{tab:datasets}
\end{table}

\subsection{Comparison Scheme}

We considered various possible competitors discussed in  \autoref{sec:RELATED WORK} as baselines.
We first selected word-embedding-based \textbf{WE-FORGE}, the state-of-the-art fake document generator for Cyber Deception.
Thus, we ignored other cyber deception and adversarial augmentation models using word embeddings. 
Instead, we chose \textbf{EDA}~\citep{eda} as a typical context-agnostic adversarial augmentation baseline.
We also compared \textbf{GPT-2} small model  (which serves as FDI's base model) to validate the advantage of the proposed mask-then-infill strategy.
We finetuned it and FDI on each training set (details in \autoref{sec:implementation}).
Finally, we ignored other controllable text generators or contextual perturbation models \citep{clare}. These methods are neither computationally efficient or show a clear advantage over the selected models on fine-grained control over concepts for this task.

\subsection{Evaluation Design}

We sampled documents from each test sets with similar lengths (e.g., 180 to 200 tokens for CS dataset) and generated their fake versions using 4 models.
We combined NLG and Cyber Deception evaluation methods to design our experiments. We collected reviews from more than 40 computer science students. Our experiments consist of \textit{Quiz-1 Detection} and \textit{Quiz-2 Evaluation}.

\paragraph{Quiz-1} We followed a similar human evaluation schema utilized in cyber deception \citep{forge,weforge} and machine-generated text detection \citep{liu2016not,van2019best,ippolito2020automatic,zellers2021turingadvice,clark2021all} to evaluate whether the fake samples can deter hackers.
Reviewers were asked to identify the original document among three fake copies generated by a single (unknown) model in each \textit{example set} (1 true + 3 fake). Each reviewer analyzed $4h$ example sets (i.e., $4h \times (1+3)$ documents) to evaluate all four models $h$ times. Finally, we computed each model's average detection accuracy and evaluation time. 

However, Quiz-1 ignores the effects of distinct generation patterns and amounts of fake content.
For example, a generated sample with minor modifications (e.g., adding or deleting a few stopwords or replacing synonyms) is less distinguishable. 
Yet, it does not protect any original document's information for the cyber deception purpose.

\paragraph{Quiz-2} 
To overcome Quiz-1's limitations, 
we designed Quiz-2 to evaluate fake samples' quality and effectiveness.
Each question set includes one known original document and four fake copies generated by four models in an unknown order.
Reviewers were asked to evaluate five metrics for the fake samples based on a 4-point Likert scale: 
(1) \textbf{fluency} of the article; 
(2) \textbf{coherency} of the article;
(3) expert knowledge (\textbf{expertise}) required to identify the article is fake;
(4) \textbf{fakeness} of the article;
and (5) the overall \textbf{preference} in the articles.

The above scores combine standard NLG metrics (fluency and coherency) and metrics we design for cyber deception. 
Specifically, fakeness indicates the amount and the effectiveness of modification applied to the original document to deceive the adversary and protect certain essential facts.  
We define four fakeness categories: \textbf{1-inadequate}, \textbf{2-marginal}, \textbf{3-moderate}, and \textbf{4-excessive}.
We do not use overlap-based metrics such as BLEU \citep{papineni2002bleu} as they are inappropriate for evaluating many realistic infills without word-level overlap \citep{ilm}.
See \autoref{sec:Questionnaire} for more details of our questionnaire.

\begin{table}
\small
\centering
\setlength{\tabcolsep}{3.2pt}
\begin{tabular}{ccccc|cc} 
\hline
\multirow{2}{*}{} & \multicolumn{4}{c|}{\# of example sets} & \multirow{2}{*}{Config} & \multirow{2}{*}{\begin{tabular}[c]{@{}c@{}}\# of fake \\ samples\end{tabular}}  \\
                  & CS  & PAT & WSJ & all                   &                         &                                                                                 \\ 
\hline
Quiz-1            & 160 & 88  & 136 & 384                   & 1 true + 3 fake                & 1152                                                                            \\
Quiz-2            & 32  & 18  & 32  & 82                    & 1 true + 4 fake                 & 328                                                                             \\
\hline
\end{tabular}
\caption{Statistics of experimented evaluation sets}
\label{tab:evaluation summary}
\end{table}

\begin{table}\setlength{\tabcolsep}{3.5pt}\small
\centering
\begin{tabular}{crcccc}
\toprule
Data    & Metric & EDA    & WEF    & GPT    & FDI    \\
\midrule
\multirow{2}{*}{CS}   & Acc $\downarrow$    & 0.93    & 0.93    & 0.65    & \textbf{0.60}    \\
   & Time $\uparrow$   & 1.53$\pm$1.0 & 2.53$\pm$1.5 & 3.31$\pm$0.3    & \textbf{3.53}$\pm$1.3  \\
\midrule
\multirow{2}{*}{PAT}  & Acc $\downarrow$    & 1.00    & 0.86    & 0.77    & \textbf{0.64}    \\
   & Time $\uparrow$   & 2.91$\pm$2.2 & 4.03$\pm$2.5 & \textbf{4.41}$\pm$2.5 & 4.17$\pm$2.0   \\
\midrule
\multirow{2}{*}{WSJ}  & Acc $\downarrow$    & 0.82    & 0.82    & \textbf{0.62}    & 0.74    \\
   & Time $\uparrow$   & 2.21$\pm$1.7 & 2.45$\pm$1.7 & \textbf{3.98}$\pm$1.3 & 3.86$\pm$1.3    \\
\midrule
\multirow{2}{*}{avg.} & Acc $\downarrow$    & 0.91    & 0.88    & 0.67    & \textbf{0.66}    \\
   & Time $\uparrow$   & 2.22$\pm$1.6 & 3.00$\pm$1.9 & \textbf{3.90}$\pm$1.5 & 3.85$\pm$1.4 \\
\bottomrule
\end{tabular}
\caption{Mean accuracy and time taken (in minutes) by participants to review one example in Quiz-1.}
\label{tab:quiz-1}
\end{table}

\subsection{Results}

\autoref{tab:evaluation summary} shows statistics of experimented evaluation sets.
Specifically, we evaluated 384 example sets in Quiz-1 (96 sets per model and 1,152 fake examples overall. For Quiz-2, we tested 82 example sets, including 328 fake samples. These samples come from the same 30 articles and their 360 fake copies. In addition, each evaluated set was evaluated by at least two students. 

\autoref{tab:quiz-1} shows the mean detection accuracy and the average time taken for participants to review one example in each scenario in Quiz-1.
Compared with other context-agnostic baselines, GPT-2 and FDI get lower accuracy and longer time. The results indicate that examining texts generated by current LMs requires more effort than a superficial judgment based on fluency-related quality aspects \citep{clark2021all}.
Although time metrics are relatively similar for these models, FDI's superiority varies across domains. It presents lower accuracy (i.e., better at misleading humans) in CS and PAT but higher in WSJ.

\begin{table}
\small
\centering
\begin{tabular}{crcccc} 
\toprule
{Data} & {Metric} & {EDA} & {WEF} & {GPT} & {FDI}\\ 
\midrule
\multirow{4}{*}{CS}   & Flu $\uparrow$   & 1.97 & 2.97 & 3.06 & \textbf{3.19}  \\
                      & Coh $\uparrow$   & 2.28 & 2.94 & 2.84 & \textbf{3.25}  \\
                      & Exp $\uparrow$   & 1.78 & 2.84 & 2.81 & \textbf{3.00}  \\
                      & Pref $\uparrow$  & 1.66 & 2.66 & 2.56 & \textbf{3.19}  \\ 
\midrule
\multirow{4}{*}{PAT}  & Flu $\uparrow$   & 1.39 & 3.22 & \textbf{3.33} & 3.28  \\
                      & Coh $\uparrow$  & 1.56 & 2.72 & 2.67 & \textbf{3.17}  \\
                      & Exp $\uparrow$   & 1.33 & 2.72 & 2.67 & \textbf{3.06}  \\
                      & Pref $\uparrow$  & 1.11 & 2.72 & 2.72 & \textbf{3.44}  \\ 
\midrule
\multirow{4}{*}{WSJ}  & Flu $\uparrow$   & 1.75 & 2.81 & \textbf{3.28} & 3.06  \\
                      & Coh $\uparrow$   & 1.78 & 2.28 & 2.63 & \textbf{2.78}  \\
                      & Exp $\uparrow$   & 1.69 & 1.84 & \textbf{2.72} & 2.59  \\
                      & Pref $\uparrow$  & 1.72 & 2.28 & 2.88 & \textbf{3.13}  \\ 
\midrule
\multirow{4}{*}{avg.} & Flu $\uparrow$   & 1.76 & 2.96 & \textbf{3.21} & 3.16  \\
                      & Coh $\uparrow$    & 1.93 & 2.63 & 2.72 & \textbf{3.05}  \\
                      & Exp $\uparrow$   & 1.65 & 2.43 & 2.74 & \textbf{2.85}  \\
                      & Pref $\uparrow$   & 1.56 & 2.52 & 2.72 & \textbf{3.22}  \\
\bottomrule
\end{tabular}
\caption{Mean scores of \textbf{flu}ency, \textbf{coh}erency, \textbf{exp}ertise, and \textbf{pref}erence in Quiz-2.}
\label{tab:quiz-2}
\end{table}

\begin{figure}[t]
  \centering
  \includegraphics[width=0.8\linewidth]{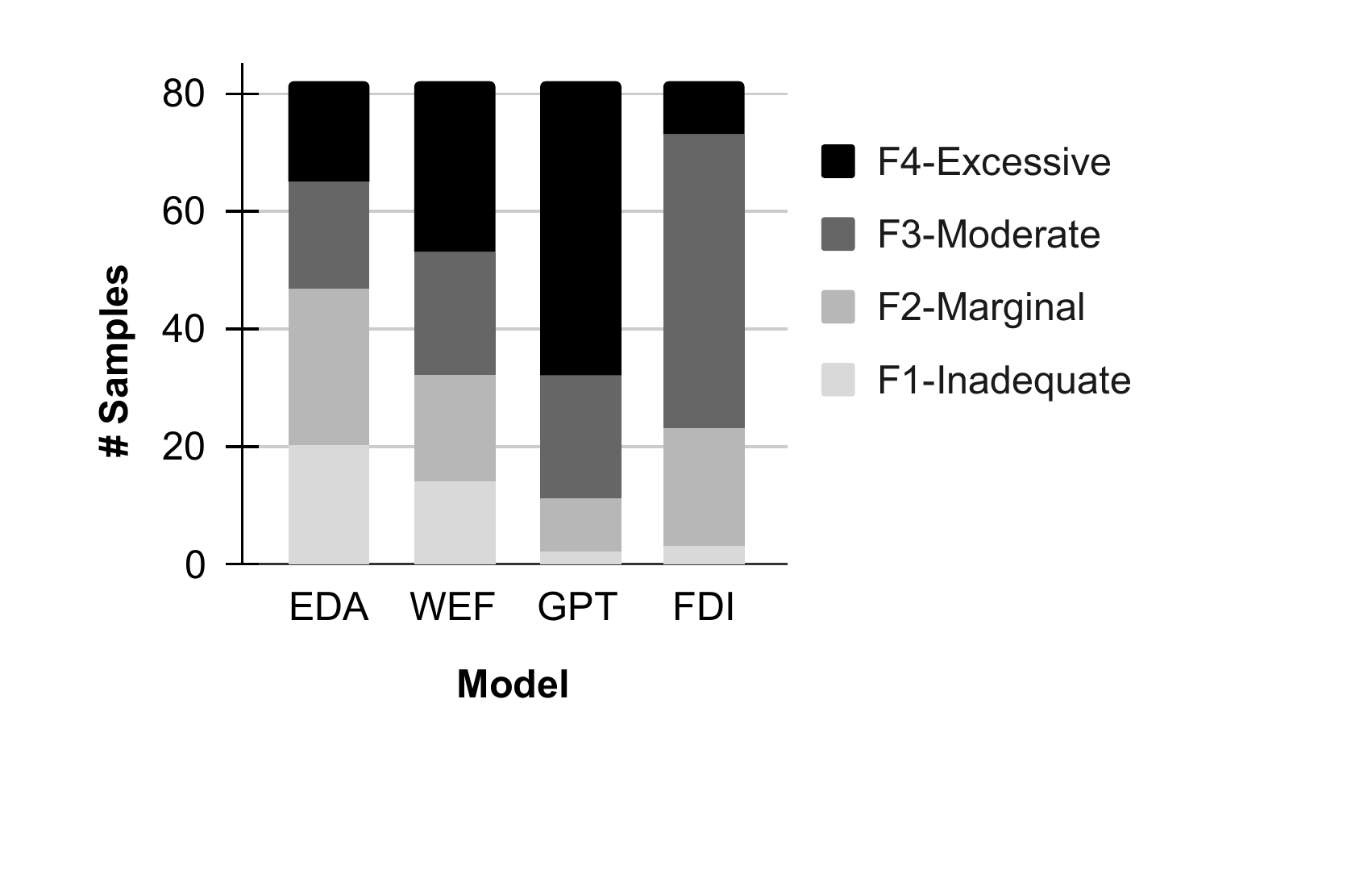}
  \caption{Distribution of the samples' fakeness. 
  Most of FDI's samples (61.0\%) have moderate fakeness.
  }
  \label{fig:fakeness}
\end{figure}

\autoref{tab:quiz-2} compiles the reviews of Quiz-2 for a more comprehensive analysis of the generated fake documents. We illustrate the fakeness metric separately in \autoref{fig:fakeness} due to its particularity (higher fakeness doesn’t mean superiority).
\autoref{tab:quiz-2} shows that EDA achieves the worst fluency and coherency due to its random perturbation strategy. 
GPT-2 generates the most fluent output with contextual knowledge in the unrestricted left-to-right manner. 
However, its output lacks  fine-grained control and gradually goes off-topic, thus affecting its coherency.
WE-FORGE and FDI preserve the article's logical and consistent relation by replacing specific snippets.  
Yet, WE-FORGE results in unstable performance due to its unigram replacement based on unbalanced word embeddings clusters.
In contrast, FDI combines improved replacement strategies and contextual features, consistently reporting superior coherency and fluency.

The expertise score refers to the level of expert knowledge required for the reviewer to identify whether the article is fake or not.
Given the low fluency and coherency, EDA's fake samples require lower expertise to be recognized.
WE-FORGE prunes out all words other than nouns because such terms are unlikely to contribute to the content of a technical document \cite{weforge}. Yet, this method hinders its outputs' diversity in a news story with fewer important nouns such as technical terms but more essential verbs.
As a result, it may generate easily identifiable fake samples such as replacing ``President Joe \underline{Biden}'' with ``President Joe \underline{Trump}''.  
Therefore, WE-FORGE is competitive with GPT-2 in CS and PAT but performs poorly in WSJ.
In contrast, GPT-2 avoids the above issues caused by replacing unigram, which also explains its superiority in accuracy and expertise score in WSJ.
FDI addresses WE-FORGE's issue by replacing $n$-grams respecting both the preceding and the following context.
Thus, its errors related to reviewers' knowledge are more subtle. 
Although we focus on technical datasets, these results suggest that FDI generalizes well in other domains.

The ideal fake samples should have moderate fakeness, neither too close nor too far away from the original text.
\autoref{fig:fakeness} illustrates that 61.0\% of FDI's generated samples have moderate fakeness, achieving the best trade-offs.
In contrast, EDA is ineffective in protecting critical information because 57.3\% of its samples have marginal or inadequate fakeness.
WE-FORGE applies more effective modification than EDA.
Yet, the near-uniform distribution of WE-FORGE's fakeness is consistent with its unstable performance.
GPT-2's samples tend to introduce excessive fakeness, substantially diverging from the original documents.

\begin{figure}[t]
  \centering
  \includegraphics[width=0.7\linewidth]{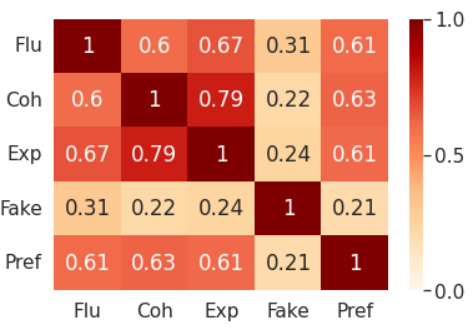}
  \caption{The Spearman correlation heatmap for fluency, coherency, expertise, fakeness, and preference scores.}
  \label{fig:correlation}
\end{figure}

In the final question of Quiz-2, we asked the reviewers to rank their favorite fake articles from score 4 to score 1.  Then we calculate each model's average results as the preference scores.
\autoref{tab:quiz-2} shows that FDI is the overall best model.
Based on the participants' feedback, various factors influence their decision-making.
For example, some reviewers like the most fluent samples, while others prefer those with realistic modification.
Therefore, we analyze the relationships between these metrics in \autoref{fig:correlation}, which illustrates that all the scores other than fakeness show strong positive correlations.
The results are as expected as we prefer fluent, coherent fake documents that require expert knowledge to identify.  
In contrast, we observe weak positive correlations between fakeness and the other metrics.

To understand the human preference in fakeness, \autoref{tab:fakeness and perference} summarizes the mean preference scores of the samples of different fakeness types.   
And it shows that the reviewers favor the samples with moderate fakeness. 
The above observation again validates a trade-off between the amount of fake content and the superiority of the fake samples.
It also indicates that fakeness is a relatively independent metric from the other evaluation metrics. Thus, it is necessary to include fakeness in the future cyber deception study.

\begin{table}[t]
\small
\centering
\begin{tabular}{lcc}
\toprule
\textbf{Fakeness}   & \textbf{Pref} & \textbf{\# count}  \\
\midrule
1-Inadequate & 1.69 & 39   \\
2-Marginal   & 2.20 & 74   \\
3-Moderate   & \textbf{2.98} & 110  \\
4-Excessive  & 2.52 & 105 \\
\bottomrule
\end{tabular}
\caption{Mean preference and the number of example sets for each fakeness type. 
}
\label{tab:fakeness and perference}
\end{table}

\subsection{Parameter Study}

Due to the extensive time and efforts associated with human-driven experiments, we used the same hyperparameters for all datasets based on evaluation results on small validation sets (details in \autoref{sec:hyperparameters}).  
A key hyperparameter is max masked rate $\gamma$, as shown in Algorithm \ref{alg:alg}.
Samples with low $\gamma$ (e.g.,$10\%$) are likely to be labeled as inadequate fakeness.
In contrast,  high $\gamma$ results in excessive fakeness and errors because the model needs to fill in more blanks given less context. 
As moderate fakeness is desired in cyber deception work, we set $\gamma=20\%$.
Yet, users can specify their preferred $\gamma$ in custom datasets.
Besides $\gamma$, many parameters provide randomness in the samples but do not significantly affect the human evaluation result.

\section{Conclusion and Future Work} 
\label{sec:CONCLUSION}
We propose a novel fake document generator, FDI, for network intrusion defense and intellectual property protection. 
FDI relies on a complete mask-then-infill process with a curated strategy for fake documents generation.
Our experiments explore ``how easily the original documents are identified'' and ``how critical information is protected'' with more fake samples and generation patterns.
FDI shows consistent superiority in generating realistic fake samples while protecting the information and deceiving the hackers.

While human evaluation remains the gold standard for evaluating various NLG applications, future work can explore automatic detection methods \cite{Grover,gltr2019,bakhtin2019real,schuster2020limitations} to alleviate human efforts. 
Besides, this work focuses on technical documents and shows generalization in news stories. Future work can also extend its applications to other critical domains, such as political science \cite{parolin2022ConfliT5,parolin2021come,hu2022conflibert,skorupa2022multi,hu2021uncertainty}.

\section{Limitations}

Due to the expensive human evaluation, we empirically selected some configurations on small validation sets. Besides, we reduced the overlaps between the reviewers to cover more samples and reduce the randomness. Although at least two reviewers evaluated each article set, the overlap was small to calculate Kappa. We were aware that evaluators might calibrate the metrics differently without training, a commonly reported issue in NLG tasks \citep{ippolito2020automatic,clark2021all}. However, pre-evaluation training on fakeness introduced bias because the reviewers may judge only based on the distinct patterns of different models (as shown in \autoref{tab:fake_sample_intro}). Thus, we didn't intervene in the evaluation. Instead, we extensively analyzed reviewers' choices in Figures \ref{fig:fakeness}, \ref{fig:correlation}, and \autoref{tab:fakeness and perference}. 
More work needs to be done by (1) designing simple but unbiased instructions to help reviewers score more consistently. (2) More overlapping experiments between reviewers to calculate Kappa.

Second, FDI is not flawless and suffers from similar weaknesses as all LMs. Text infilling models may generate repetitive text, incomplete words, or unmatched parenthesis, resulting in a high infilling failure rate \citep{shen2020blank}. Therefore, we designed several heuristic steps in Lines \ref{alg:concatDet}, \ref{alg:getsent}, and \ref{alg:merge_close_concepts} of Algorithm \ref{alg:alg} to simplify the infilling tasks and reduce errors. We believe a more powerful LM, such as GPT-3 \cite{brown2020language}, can improve the performance further. Besides, GPT-2 is originally pretrained for left-to-right text generation. Some alternative LMs, such as T-5 \cite{t5} and BART \cite{bart}, have already learned elementary text-infilling tasks during the pretraining. Future work should also explore how these models perform in our framework.

Finally, we designed a simple penalized decoding strategy based on Top-$p$\% schema to encourage diverse fake generations. Yet, it also generated errors like other constrained decoding methods. 
Future work should optimize the decoding algorithm and post-processing methods.

\section{Ethical Considerations} 
\label{sec:Ethical Considerations}
We  acknowledge that similar mechanisms may be abused to generate disinformation, such as fake news~\citep{Grover}.
Besides, language models have been shown to encode biases from the training data \citep{barbera2021automated}.
Thus, we remove controversial and sensitive news samples to mitigate these issues during our evaluation.
With the rapid evolution of Cyber Deception and NLG technologies, we believe this work creates more value than risks on balance.

\section*{Acknowledgments}
The research reported herein was supported in part by NSF awards DMS-1737978, DGE-2039542, OAC-1828467, OAC-1931541, and DGE-1906630, ONR awards N00014-17-1-2995 and N00014-20-1-2738, Army Research Office Contract No. W911NF2110032 and IBM faculty award (Research). 
We sincerely thank all the participants in the questionnaire for their valuable contributions.

\bibliography{reference}

\begin{thebibliography}{77}
\expandafter\ifx\csname natexlab\endcsname\relax\def\natexlab#1{#1}\fi

\bibitem[{Abdibayev et~al.(2021)Abdibayev, Chen, Chen, Poluru, and
  Subrahmanian}]{weforge}
Almas Abdibayev, Dongkai Chen, Haipeng Chen, Deepti Poluru, and V.~S.
  Subrahmanian. 2021.
\newblock \href {https://doi.org/10.1145/3418289} {Using word embeddings to
  deter intellectual property theft through automated generation of fake
  documents}.
\newblock \emph{ACM Trans. Manage. Inf. Syst.}, 12(2).

\bibitem[{Akbar et~al.(2022)Akbar, Halim, Hu, Singhal, Khan, and
  Thuraisingham}]{akbar2022knowledge}
Khandakar~Ashrafi Akbar, Sadaf~Md Halim, Yibo Hu, Anoop Singhal, Latifur Khan,
  and Bhavani Thuraisingham. 2022.
\newblock Knowledge mining in cybersecurity: From attack to defense.
\newblock In \emph{IFIP Annual Conference on Data and Applications Security and
  Privacy}, pages 110--122. Springer.

\bibitem[{Alzantot et~al.(2018)Alzantot, Sharma, Elgohary, Ho, Srivastava, and
  Chang}]{alzantot2018generating}
Moustafa Alzantot, Yash Sharma, Ahmed Elgohary, Bo-Jhang Ho, Mani Srivastava,
  and Kai-Wei Chang. 2018.
\newblock Generating natural language adversarial examples.
\newblock In \emph{Proceedings of the 2018 Conference on Empirical Methods in
  Natural Language Processing}, pages 2890--2896.

\bibitem[{Bakhtin et~al.(2019)Bakhtin, Gross, Ott, Deng, Ranzato, and
  Szlam}]{bakhtin2019real}
Anton Bakhtin, Sam Gross, Myle Ott, Yuntian Deng, Marc'Aurelio Ranzato, and
  Arthur Szlam. 2019.
\newblock Real or fake? learning to discriminate machine from human generated
  text.

\bibitem[{Barber{\'a} et~al.(2021)Barber{\'a}, Boydstun, Linn, McMahon, and
  Nagler}]{barbera2021automated}
Pablo Barber{\'a}, Amber~E Boydstun, Suzanna Linn, Ryan McMahon, and Jonathan
  Nagler. 2021.
\newblock Automated text classification of news articles: A practical guide.
\newblock \emph{Political Analysis}, 29(1):19--42.

\bibitem[{Biewald(2020)}]{wandb}
Lukas Biewald. 2020.
\newblock \href {https://www.wandb.com/} {Experiment tracking with weights and
  biases}.
\newblock Software available from wandb.com.

\bibitem[{Bowen et~al.(2009)Bowen, Hershkop, Keromytis, and
  Stolfo}]{bowen2009baiting}
Brian~M Bowen, Shlomo Hershkop, Angelos~D Keromytis, and Salvatore~J Stolfo.
  2009.
\newblock Baiting inside attackers using decoy documents.
\newblock In \emph{International Conference on Security and Privacy in
  Communication Systems}, pages 51--70. Springer.

\bibitem[{Brown et~al.(2020)Brown, Mann, Ryder, Subbiah, Kaplan, Dhariwal,
  Neelakantan, Shyam, Sastry, Askell et~al.}]{brown2020language}
Tom~B Brown, Benjamin Mann, Nick Ryder, Melanie Subbiah, Jared Kaplan, Prafulla
  Dhariwal, Arvind Neelakantan, Pranav Shyam, Girish Sastry, Amanda Askell,
  et~al. 2020.
\newblock Language models are few-shot learners.
\newblock \emph{arXiv preprint arXiv:2005.14165}.

\bibitem[{Chakraborty et~al.(2021)Chakraborty, Jajodia, Katz, Picariello,
  Sperli, and Subrahmanian}]{forge}
Tanmoy Chakraborty, Sushil Jajodia, Jonathan Katz, Antonio Picariello,
  Giancarlo Sperli, and V.~S. Subrahmanian. 2021.
\newblock \href {https://doi.org/10.1109/TDSC.2019.2898661} {A fake online
  repository generation engine for cyber deception}.
\newblock \emph{IEEE Transactions on Dependable and Secure Computing},
  18(2):518--533.

\bibitem[{Clark et~al.(2021)Clark, August, Serrano, Haduong, Gururangan, and
  Smith}]{clark2021all}
Elizabeth Clark, Tal August, Sofia Serrano, Nikita Haduong, Suchin Gururangan,
  and Noah~A Smith. 2021.
\newblock All that’s ‘human’is not gold: Evaluating human evaluation of
  generated text.
\newblock In \emph{Proceedings of the 59th Annual Meeting of the Association
  for Computational Linguistics and the 11th International Joint Conference on
  Natural Language Processing (Volume 1: Long Papers)}, pages 7282--7296.

\bibitem[{Clark et~al.(2018)Clark, Ji, and Smith}]{clark2018neural}
Elizabeth Clark, Yangfeng Ji, and Noah~A Smith. 2018.
\newblock Neural text generation in stories using entity representations as
  context.
\newblock In \emph{Association for Computational Linguistics: Human Language
  Technologies}.

\bibitem[{Clayton(2017)}]{cyber}
Jay Clayton. 2017.
\newblock \href
  {https://www.sec.gov/news/public-statement/statement-clayton-2017-09-20}
  {Statement on cybersecurity}.
\newblock Accessed: 2021-11-10.

\bibitem[{Das and Verma(2020)}]{das2020can}
Avisha Das and Rakesh~M Verma. 2020.
\newblock Can machines tell stories? a comparative study of deep neural
  language models and metrics.
\newblock \emph{IEEE Access}, 8:181258--181292.

\bibitem[{Dathathri et~al.(2020)Dathathri, Madotto, Lan, Hung, Frank, Molino,
  Yosinski, and Liu}]{pplm}
Sumanth Dathathri, Andrea Madotto, Janice Lan, Jane Hung, Eric Frank, Piero
  Molino, Jason Yosinski, and Rosanne Liu. 2020.
\newblock \href {https://openreview.net/forum?id=H1edEyBKDS} {Plug and play
  language models: A simple approach to controlled text generation}.
\newblock In \emph{International Conference on Learning Representations}.

\bibitem[{Devlin et~al.(2018)Devlin, Chang, Lee, and Toutanova}]{bert}
Jacob Devlin, Ming-Wei Chang, Kenton Lee, and Kristina Toutanova. 2018.
\newblock Bert: Pre-training of deep bidirectional transformers for language
  understanding.
\newblock \emph{arXiv preprint arXiv:1810.04805}.

\bibitem[{Donahue et~al.(2020)Donahue, Lee, and Liang}]{ilm}
Chris Donahue, Mina Lee, and Percy Liang. 2020.
\newblock Enabling language models to fill in the blanks.
\newblock In \emph{Proceedings of the 58th Annual Meeting of the Association
  for Computational Linguistics}.

\bibitem[{Fan et~al.(2018)Fan, Lewis, and Dauphin}]{fan2018hierarchical}
A.~Fan, M.~Lewis, and Y.~Dauphin. 2018.
\newblock Hierarchical neural story generation.
\newblock \emph{arXiv preprint arXiv:1805.04833}.

\bibitem[{Fan et~al.(2019)Fan, Lewis, and Dauphin}]{fan2019strategies}
Angela Fan, Mike Lewis, and Yann Dauphin. 2019.
\newblock Strategies for structuring story generation.
\newblock In \emph{Proceedings of the 57th Annual Meeting of the Association
  for Computational Linguistics}, pages 2650--2660.

\bibitem[{Fedus et~al.(2018)Fedus, Goodfellow, and Dai}]{fedus2018maskgan}
William Fedus, Ian Goodfellow, and Andrew~M Dai. 2018.
\newblock Maskgan: Better text generation via filling in the \_.
\newblock In \emph{International Conference on Learning Representations}.

\bibitem[{Garg and Ramakrishnan(2020)}]{garg2020bae}
Siddhant Garg and Goutham Ramakrishnan. 2020.
\newblock Bae: Bert-based adversarial examples for text classification.
\newblock In \emph{Proceedings of the 2020 Conference on Empirical Methods in
  Natural Language Processing (EMNLP)}, pages 6174--6181.

\bibitem[{Gehrmann et~al.(2019)Gehrmann, Strobelt, and Rush}]{gltr2019}
Sebastian Gehrmann, Hendrik Strobelt, and Alexander Rush. 2019.
\newblock \href {https://doi.org/10.18653/v1/P19-3019} {{GLTR}: Statistical
  detection and visualization of generated text}.
\newblock In \emph{Proceedings of the 57th Annual Meeting of the Association
  for Computational Linguistics: System Demonstrations}, pages 111--116,
  Florence, Italy. Association for Computational Linguistics.

\bibitem[{Goldfarb-Tarrant et~al.(2020)Goldfarb-Tarrant, Chakrabarty,
  Weischedel, and Peng}]{goldfarb2020content}
Seraphina Goldfarb-Tarrant, Tuhin Chakrabarty, Ralph Weischedel, and Nanyun
  Peng. 2020.
\newblock Content planning for neural story generation with aristotelian
  rescoring.
\newblock In \emph{Proceedings of the 2020 Conference on Empirical Methods in
  Natural Language Processing (EMNLP)}, pages 4319--4338.

\bibitem[{Holtzman et~al.(2019)Holtzman, Buys, Du, Forbes, and
  Choi}]{holtzman2019curious}
Ari Holtzman, Jan Buys, Li~Du, Maxwell Forbes, and Yejin Choi. 2019.
\newblock The curious case of neural text degeneration.
\newblock In \emph{International Conference on Learning Representations}.

\bibitem[{Hu et~al.(2022)Hu, Hosseini, Parolin, Osorio, Khan, Brandt, and
  D’Orazio}]{hu2022conflibert}
Yibo Hu, MohammadSaleh Hosseini, Erick~Skorupa Parolin, Javier Osorio, Latifur
  Khan, Patrick Brandt, and Vito D’Orazio. 2022.
\newblock Conflibert: A pre-trained language model for political conflict and
  violence.
\newblock In \emph{Proceedings of the 2022 Conference of the North American
  Chapter of the Association for Computational Linguistics: Human Language
  Technologies}, pages 5469--5482.

\bibitem[{Hu and Khan(2021)}]{hu2021uncertainty}
Yibo Hu and Latifur Khan. 2021.
\newblock Uncertainty-aware reliable text classification.
\newblock In \emph{Proceedings of the 27th ACM SIGKDD Conference on Knowledge
  Discovery \& Data Mining}, pages 628--636.

\bibitem[{Ippolito et~al.(2019)Ippolito, Grangier, Callison-Burch, and
  Eck}]{ippolito2019unsupervised}
D.~Ippolito, D.~Grangier, C.~Callison-Burch, and D.~Eck. 2019.
\newblock Unsupervised hierarchical story infilling.
\newblock In \emph{NAACL Workshop on Narrative Understanding}, pages 37--43.

\bibitem[{Ippolito et~al.(2020{\natexlab{a}})Ippolito, Duckworth,
  Callison-Burch, and Eck}]{ippolito2020automatic}
Daphne Ippolito, Daniel Duckworth, Chris Callison-Burch, and Douglas Eck.
  2020{\natexlab{a}}.
\newblock Automatic detection of generated text is easiest when humans are
  fooled.
\newblock In \emph{Proceedings of the 58th Annual Meeting of the Association
  for Computational Linguistics}, pages 1808--1822.

\bibitem[{Ippolito et~al.(2020{\natexlab{b}})Ippolito, Grangier, Eck, and
  Callison-Burch}]{Ippolito2020TowardBS}
Daphne Ippolito, David Grangier, D.~Eck, and Chris Callison-Burch.
  2020{\natexlab{b}}.
\newblock Toward better storylines with sentence-level language models.
\newblock In \emph{ACL}.

\bibitem[{Jia et~al.(2019)Jia, Raghunathan, G{\"o}ksel, and
  Liang}]{jia2019certified}
Robin Jia, Aditi Raghunathan, Kerem G{\"o}ksel, and Percy Liang. 2019.
\newblock Certified robustness to adversarial word substitutions.
\newblock In \emph{Proceedings of the 2019 Conference on Empirical Methods in
  Natural Language Processing and the 9th International Joint Conference on
  Natural Language Processing (EMNLP-IJCNLP)}, pages 4129--4142.

\bibitem[{Karuna et~al.(2021)Karuna, Purohit, Jajodia, Ganesan, and
  Uzuner}]{fake_karuna2021}
Prakruthi Karuna, Hemant Purohit, Sushil Jajodia, Rajesh Ganesan, and Ozlem
  Uzuner. 2021.
\newblock \href {https://doi.org/10.1109/JSYST.2020.2980177} {Fake document
  generation for cyber deception by manipulating text comprehensibility}.
\newblock \emph{IEEE Systems Journal}, 15(1):835--845.

\bibitem[{Keskar et~al.(2019)Keskar, McCann, Varshney, Xiong, and
  Socher}]{ctrl}
Nitish~Shirish Keskar, Bryan McCann, Lav Varshney, Caiming Xiong, and Richard
  Socher. 2019.
\newblock {CTRL - A Conditional Transformer Language Model for Controllable
  Generation}.
\newblock \emph{arXiv preprint arXiv:1909.05858}.

\bibitem[{Kingma and Ba(2014)}]{adam_optimizer}
Diederik~P Kingma and Jimmy Ba. 2014.
\newblock Adam: A method for stochastic optimization.
\newblock \emph{arXiv preprint arXiv:1412.6980}.

\bibitem[{Krause et~al.(2020)Krause, Gotmare, McCann, Keskar, Joty, Socher, and
  Rajani}]{KrauseGeDi2020}
Ben Krause, Akhilesh~Deepak Gotmare, Bryan McCann, Nitish~Shirish Keskar,
  Shafiq Joty, Richard Socher, and Nazneen~Fatema Rajani. 2020.
\newblock {GeDi: Generative Discriminator Guided Sequence Generation}.
\newblock \emph{arXiv preprint arXiv:2009.06367}.

\bibitem[{Lewis et~al.(2019)Lewis, Liu, Goyal, Ghazvininejad, Mohamed, Levy,
  Stoyanov, and Zettlemoyer}]{bart}
Mike Lewis, Yinhan Liu, Naman Goyal, Marjan Ghazvininejad, Abdelrahman Mohamed,
  Omer Levy, Ves Stoyanov, and Luke Zettlemoyer. 2019.
\newblock Bart: Denoising sequence-to-sequence pre-training for natural
  language generation, translation, and comprehension.
\newblock \emph{arXiv preprint arXiv:1910.13461}.

\bibitem[{Li et~al.(2021)Li, Zhang, Peng, Chen, Brockett, Sun, and
  Dolan}]{clare}
Dianqi Li, Yizhe Zhang, Hao Peng, Liqun Chen, Chris Brockett, Ming-Ting Sun,
  and William~B Dolan. 2021.
\newblock Contextualized perturbation for textual adversarial attack.
\newblock In \emph{Proceedings of the 2021 Conference of the North American
  Chapter of the Association for Computational Linguistics: Human Language
  Technologies}, pages 5053--5069.

\bibitem[{Li et~al.(2020)Li, Ma, Guo, Xue, and Qiu}]{li2020bert}
Linyang Li, Ruotian Ma, Qipeng Guo, Xiangyang Xue, and Xipeng Qiu. 2020.
\newblock Bert-attack: Adversarial attack against bert using bert.
\newblock In \emph{Proceedings of the 2020 Conference on Empirical Methods in
  Natural Language Processing (EMNLP)}, pages 6193--6202.

\bibitem[{Liu et~al.(2021)Liu, Sap, Lu, Swayamdipta, Bhagavatula, Smith, and
  Choi}]{liu2021dexperts}
Alisa Liu, Maarten Sap, Ximing Lu, Swabha Swayamdipta, Chandra Bhagavatula,
  Noah~A Smith, and Yejin Choi. 2021.
\newblock Dexperts: Decoding-time controlled text generation with experts and
  anti-experts.
\newblock In \emph{Proceedings of the 59th Annual Meeting of the Association
  for Computational Linguistics and the 11th International Joint Conference on
  Natural Language Processing (Volume 1: Long Papers)}, pages 6691--6706.

\bibitem[{Liu et~al.(2016)Liu, Lowe, Serban, Noseworthy, Charlin, and
  Pineau}]{liu2016not}
Chia-Wei Liu, Ryan Lowe, Iulian~Vlad Serban, Mike Noseworthy, Laurent Charlin,
  and Joelle Pineau. 2016.
\newblock How not to evaluate your dialogue system: An empirical study of
  unsupervised evaluation metrics for dialogue response generation.
\newblock In \emph{Proceedings of the 2016 Conference on Empirical Methods in
  Natural Language Processing}, pages 2122--2132.

\bibitem[{Liu et~al.(2019)Liu, Fu, Liu, and Lv}]{liu-etal-2019-tigs}
Dayiheng Liu, Jie Fu, Pengfei Liu, and Jiancheng Lv. 2019.
\newblock \href {https://doi.org/10.18653/v1/P19-1406} {{TIGS}: An inference
  algorithm for text infilling with gradient search}.
\newblock In \emph{Proceedings of the 57th Annual Meeting of the Association
  for Computational Linguistics}, pages 4146--4156, Florence, Italy.
  Association for Computational Linguistics.

\bibitem[{Masud et~al.(2007)Masud, Khan, and Thuraisingham}]{masud2007hybrid}
Mohammad~M Masud, Latifur Khan, and Bhavani Thuraisingham. 2007.
\newblock A hybrid model to detect malicious executables.
\newblock In \emph{2007 IEEE International Conference on Communications}, pages
  1443--1448. IEEE.

\bibitem[{Mikolov et~al.(2013)Mikolov, Chen, Corrado, and
  Dean}]{Mikolov2013EfficientEO}
Tomas Mikolov, Kai Chen, Gregory~S. Corrado, and Jeffrey Dean. 2013.
\newblock Efficient estimation of word representations in vector space.
\newblock In \emph{ICLR}.

\bibitem[{Mikolov et~al.(2018)Mikolov, Grave, Bojanowski, Puhrsch, and
  Joulin}]{mikolov2018advances}
Tomas Mikolov, Edouard Grave, Piotr Bojanowski, Christian Puhrsch, and Armand
  Joulin. 2018.
\newblock Advances in pre-training distributed word representations.
\newblock In \emph{Proceedings of the International Conference on Language
  Resources and Evaluation (LREC 2018)}.

\bibitem[{Morris et~al.(2020)Morris, Lifland, Yoo, Grigsby, Jin, and
  Qi}]{Morris2020TextAttackAF}
John~X. Morris, Eli Lifland, Jin~Yong Yoo, Jake Grigsby, Di~Jin, and Yanjun Qi.
  2020.
\newblock Textattack: A framework for adversarial attacks, data augmentation,
  and adversarial training in nlp.
\newblock In \emph{EMNLP}.

\bibitem[{Papineni et~al.(2002)Papineni, Roukos, Ward, and
  Zhu}]{papineni2002bleu}
Kishore Papineni, Salim Roukos, Todd Ward, and Wei-Jing Zhu. 2002.
\newblock Bleu: a method for automatic evaluation of machine translation.
\newblock In \emph{Proceedings of the 40th annual meeting of the Association
  for Computational Linguistics}, pages 311--318.

\bibitem[{Parolin et~al.(2022)Parolin, Hu, Khan, Brandt, Osorio, and
  D’Orazio}]{parolin2022ConfliT5}
Erick~Skorupa Parolin, Yibo Hu, Latifur Khan, Patrick~T Brandt, Javier Osorio,
  and Vito D’Orazio. 2022.
\newblock Confli-t5: An autoprompt pipeline for conflict related text
  augmentation.
\newblock In \emph{2022 IEEE International Conference on Big Data (Big Data)}.
  IEEE.

\bibitem[{Parolin et~al.(2021)Parolin, Hu, Khan, Osorio, Brandt, and
  D’Orazio}]{parolin2021come}
Erick~Skorupa Parolin, Yibo Hu, Latifur Khan, Javier Osorio, Patrick~T Brandt,
  and Vito D’Orazio. 2021.
\newblock Come-ke: A new transformers based approach for knowledge extraction
  in conflict and mediation domain.
\newblock In \emph{2021 IEEE International Conference on Big Data (Big Data)},
  pages 1449--1459. IEEE.

\bibitem[{Pennington et~al.(2014)Pennington, Socher, and
  Manning}]{pennington2014glove}
Jeffrey Pennington, Richard Socher, and Christopher~D Manning. 2014.
\newblock Glove: Global vectors for word representation.
\newblock In \emph{Proceedings of the 2014 conference on empirical methods in
  natural language processing (EMNLP)}, pages 1532--1543.

\bibitem[{Radford et~al.(2019)Radford, Wu, Child, Luan, Amodei, and
  Sutskever}]{gpt2}
Alec Radford, Jeff Wu, R.~Child, David Luan, Dario Amodei, and Ilya Sutskever.
  2019.
\newblock Language models are unsupervised multitask learners.
\newblock \emph{OpenAI blog}, 1(8):9.

\bibitem[{Raffel et~al.(2020)Raffel, Shazeer, Roberts, Lee, Narang, Matena,
  Zhou, Li, Liu et~al.}]{t5}
Colin Raffel, Noam Shazeer, Adam Roberts, Katherine Lee, Sharan Narang, Michael
  Matena, Yanqi Zhou, Wei Li, Peter~J Liu, et~al. 2020.
\newblock Exploring the limits of transfer learning with a unified text-to-text
  transformer.
\newblock \emph{J. Mach. Learn. Res.}, 21(140):1--67.

\bibitem[{Ranade et~al.(2021)Ranade, Piplai, Mittal, Joshi, and
  Finin}]{fake_gpt}
Priyanka Ranade, Aritran Piplai, Sudip Mittal, Anupam Joshi, and Tim Finin.
  2021.
\newblock Generating fake cyber threat intelligence using transformer-based
  models.
\newblock \emph{arXiv preprint arXiv:2102.04351}.

\bibitem[{Rashkin et~al.(2020)Rashkin, Celikyilmaz, Choi, and
  Gao}]{rashkin2020plotmachines}
Hannah Rashkin, Asli Celikyilmaz, Yejin Choi, and Jianfeng Gao. 2020.
\newblock Plotmachines: Outline-conditioned generation with dynamic plot state
  tracking.
\newblock \emph{arXiv preprint arXiv:2004.14967}.

\bibitem[{Ren et~al.(2019)Ren, Deng, He, and Che}]{ren2019generating}
Shuhuai Ren, Yihe Deng, Kun He, and Wanxiang Che. 2019.
\newblock Generating natural language adversarial examples through probability
  weighted word saliency.
\newblock In \emph{Proceedings of the 57th Annual Meeting of the Association
  for Computational Linguistics}, pages 1085--1097.

\bibitem[{Ribeiro et~al.(2020)Ribeiro, Wu, Guestrin, and
  Singh}]{Ribeiro2020BeyondAB}
Marco~T{\'u}lio Ribeiro, Tongshuang~(Sherry) Wu, Carlos Guestrin, and Sameer
  Singh. 2020.
\newblock Beyond accuracy: Behavioral testing of nlp models with checklist.
\newblock In \emph{ACL}.

\bibitem[{Rose et~al.(2010)Rose, Engel, Cramer, and Cowley}]{rake}
Stuart Rose, Dave Engel, Nick Cramer, and Wendy Cowley. 2010.
\newblock Automatic keyword extraction from individual documents.
\newblock \emph{Text mining: applications and theory}, 1:1--20.

\bibitem[{Schuster et~al.(2020)Schuster, Schuster, Shah, and
  Barzilay}]{schuster2020limitations}
Tal Schuster, Roei Schuster, Darsh~J Shah, and Regina Barzilay. 2020.
\newblock The limitations of stylometry for detecting machine-generated fake
  news.
\newblock \emph{Computational Linguistics}, 46(2):499--510.

\bibitem[{Shen et~al.(2020)Shen, Quach, Barzilay, and Jaakkola}]{shen2020blank}
Tianxiao Shen, Victor Quach, Regina Barzilay, and Tommi Jaakkola. 2020.
\newblock {Blank Language Models}.
\newblock In \emph{Proceedings of the 2020 Conference on Empirical Methods in
  Natural Language Processing}, Online. Association for Computational
  Linguistics.

\bibitem[{Skorupa~Parolin et~al.(2022)Skorupa~Parolin, Hosseini, Hu, Khan,
  Brandt, Osorio, and D'Orazio}]{skorupa2022multi}
Erick Skorupa~Parolin, MohammadSaleh Hosseini, Yibo Hu, Latifur Khan, Patrick~T
  Brandt, Javier Osorio, and Vito D'Orazio. 2022.
\newblock Multi-coped: A multilingual multi-task approach for coding political
  event data on conflict and mediation domain.
\newblock In \emph{Proceedings of the 2022 AAAI/ACM Conference on AI, Ethics,
  and Society}, pages 700--711.

\bibitem[{Tan et~al.(2021)Tan, Yang, Al-Shedivat, Xing, and
  Hu}]{Tan2021ProgressiveGO}
Bowen Tan, Zichao Yang, Maruan Al-Shedivat, E.~Xing, and Zhiting Hu. 2021.
\newblock Progressive generation of long text with pretrained language models.
\newblock In \emph{NAACL}.

\bibitem[{Taylor(1953)}]{taylor1953cloze}
Wilson~L Taylor. 1953.
\newblock “cloze procedure”: A new tool for measuring readability.
\newblock \emph{Journalism quarterly}, 30(4):415--433.

\bibitem[{Tu et~al.(2008)Tu, Li, Yen, Thuraisingham, and Khan}]{tu2008secure}
Manghui Tu, Peng Li, I-Ling Yen, Bhavani~M Thuraisingham, and Latifur Khan.
  2008.
\newblock Secure data objects replication in data grid.
\newblock \emph{IEEE Transactions on dependable and secure computing},
  7(1):50--64.

\bibitem[{Van Der~Lee et~al.(2019)Van Der~Lee, Gatt, Van~Miltenburg, Wubben,
  and Krahmer}]{van2019best}
Chris Van Der~Lee, Albert Gatt, Emiel Van~Miltenburg, Sander Wubben, and Emiel
  Krahmer. 2019.
\newblock Best practices for the human evaluation of automatically generated
  text.
\newblock In \emph{Proceedings of the 12th International Conference on Natural
  Language Generation}, pages 355--368.

\bibitem[{Voris et~al.(2012)Voris, Boggs, and Stolfo}]{voris2012lost}
Jonathan Voris, Nathaniel Boggs, and Salvatore~J Stolfo. 2012.
\newblock Lost in translation: Improving decoy documents via automated
  translation.
\newblock In \emph{2012 IEEE Symposium on Security and Privacy Workshops},
  pages 129--133. IEEE.

\bibitem[{Wang et~al.(2013)Wang, Li, Tan, and Wang}]{Wang2013generation}
Lei Wang, Chenglong Li, QingFeng Tan, and XueBin Wang. 2013.
\newblock Generation and distribution of decoy document system.
\newblock In \emph{International Conference on Trustworthy Computing and
  Services}, pages 123--129. Springer.

\bibitem[{Wei and Zou(2019)}]{eda}
Jason Wei and Kai Zou. 2019.
\newblock Eda: Easy data augmentation techniques for boosting performance on
  text classification tasks.
\newblock In \emph{Proceedings of the 2019 Conference on Empirical Methods in
  Natural Language Processing and the 9th International Joint Conference on
  Natural Language Processing (EMNLP-IJCNLP)}, pages 6382--6388.

\bibitem[{White and Thompson(2006)}]{white2006using}
Jonathan White and Dale Thompson. 2006.
\newblock Using synthetic decoys to digitally watermark personally-identifying
  data and to promote data security.
\newblock In \emph{Security and Management}, pages 91--99. Citeseer.

\bibitem[{Whitham(2013)}]{whitham2013automating}
Ben Whitham. 2013.
\newblock Automating the generation of fake documents to detect network
  intruders.
\newblock \emph{International Journal of Cyber-Security and Digital Forensics},
  2(1):103.

\bibitem[{Whitham(2017)}]{whitham2017automating}
Ben Whitham. 2017.
\newblock Automating the generation of enticing text content for
  high-interaction honeyfiles.
\newblock In \emph{Proceedings of the 50th Hawaii International Conference on
  System Sciences}.

\bibitem[{Wolf et~al.(2020)Wolf, Debut, Sanh, Chaumond, Delangue, Moi, Cistac,
  Rault, Louf, Funtowicz, Davison, Shleifer, von Platen, Ma, Jernite, Plu, Xu,
  Scao, Gugger, Drame, Lhoest, and Rush}]{huggingface}
Thomas Wolf, Lysandre Debut, Victor Sanh, Julien Chaumond, Clement Delangue,
  Anthony Moi, Pierric Cistac, Tim Rault, Rémi Louf, Morgan Funtowicz, Joe
  Davison, Sam Shleifer, Patrick von Platen, Clara Ma, Yacine Jernite, Julien
  Plu, Canwen Xu, Teven~Le Scao, Sylvain Gugger, Mariama Drame, Quentin Lhoest,
  and Alexander~M. Rush. 2020.
\newblock \href {https://www.aclweb.org/anthology/2020.emnlp-demos.6}
  {Transformers: State-of-the-art natural language processing}.
\newblock In \emph{Proceedings of the 2020 Conference on Empirical Methods in
  Natural Language Processing: System Demonstrations}, pages 38--45, Online.
  Association for Computational Linguistics.

\bibitem[{Wu et~al.(2021)Wu, Ribeiro, Heer, and Weld}]{wu2021polyjuice}
Tongshuang Wu, Marco~Tulio Ribeiro, Jeffrey Heer, and Daniel~S Weld. 2021.
\newblock Polyjuice: Generating counterfactuals for explaining, evaluating, and
  improving models.
\newblock In \emph{Proceedings of the 59th Annual Meeting of the Association
  for Computational Linguistics}.

\bibitem[{Yang and Klein(2021)}]{yang2021fudge}
Kevin Yang and Dan Klein. 2021.
\newblock Fudge: Controlled text generation with future discriminators.
\newblock In \emph{Proceedings of the 2021 Conference of the North American
  Chapter of the Association for Computational Linguistics: Human Language
  Technologies}, pages 3511--3535.

\bibitem[{Yang et~al.(2019)Yang, Dai, Yang, Carbonell, Salakhutdinov, and
  Le}]{xlnet}
Zhilin Yang, Zihang Dai, Yiming Yang, Jaime Carbonell, Russ~R Salakhutdinov,
  and Quoc~V Le. 2019.
\newblock Xlnet: Generalized autoregressive pretraining for language
  understanding.
\newblock \emph{Advances in neural information processing systems}, 32.

\bibitem[{Yao et~al.(2019)Yao, Peng, Weischedel, Knight, Zhao, and
  Yan}]{peng2019plan}
L.~Yao, N.~Peng, R.~Weischedel, K.~Knight, D.~Zhao, and R.~Yan. 2019.
\newblock Plan-and-write: Towards better automatic storytelling.
\newblock In \emph{Association for the Advancement of Artificial Intelligence
  (AAAI)}.

\bibitem[{Yuill et~al.(2004)Yuill, Zappe, Denning, and
  Feer}]{yuill2004honeyfiles}
Jim Yuill, Mike Zappe, Dorothy Denning, and Fred Feer. 2004.
\newblock Honeyfiles: deceptive files for intrusion detection.
\newblock In \emph{Proceedings from the Fifth Annual IEEE SMC Information
  Assurance Workshop, 2004.}, pages 116--122. IEEE.

\bibitem[{Zaidi et~al.(2019)Zaidi, Cohn, and Haffari}]{zaidi2019decoding}
Najam Zaidi, Trevor Cohn, and Gholamreza Haffari. 2019.
\newblock Decoding as dynamic programming for recurrent autoregressive models.
\newblock In \emph{International Conference on Learning Representations}.

\bibitem[{Zellers et~al.(2021)Zellers, Holtzman, Clark, Qin, Farhadi, and
  Choi}]{zellers2021turingadvice}
Rowan Zellers, Ari Holtzman, Elizabeth Clark, Lianhui Qin, Ali Farhadi, and
  Yejin Choi. 2021.
\newblock Turingadvice: A generative and dynamic evaluation of language use.
\newblock In \emph{Proceedings of the 2021 Conference of the North American
  Chapter of the Association for Computational Linguistics: Human Language
  Technologies}, pages 4856--4880.

\bibitem[{Zellers et~al.(2019)Zellers, Holtzman, Rashkin, Bisk, Farhadi,
  Roesner, and Choi}]{Grover}
Rowan Zellers, Ari Holtzman, Hannah Rashkin, Yonatan Bisk, Ali Farhadi,
  Franziska Roesner, and Yejin Choi. 2019.
\newblock Defending against neural fake news.
\newblock In \emph{Advances in Neural Information Processing Systems 32}.

\bibitem[{Zhu et~al.(2019)Zhu, Hu, and Xing}]{Zhu2019TextI}
Wanrong Zhu, Zhiting Hu, and E.~Xing. 2019.
\newblock Text infilling.
\newblock \emph{ArXiv}, abs/1901.00158.

\end{thebibliography}
\bibliographystyle{acl_natbib}

\clearpage

\appendix

\section{Implementation}\label{sec:implementation}

We used the implementation of EDA from~\citep{Morris2020TextAttackAF}. 
We set the word swapping rate to 20\%.
For WE-FORGE, we learnt word-embeddings per dataset using Word2Vec~\citep{Mikolov2013EfficientEO}.
Based on silhouette scores and empirical observations on the validation sets, we set $k=100$ for K-means and the number of Concept-Importance-Bins as 5 for all datasets. 

For GPT-2 and FDI,
we implemented the models with Huggingface API~\citep{huggingface} and monitored the training with Wandb~\citep{wandb}.
We used Adam optimizer~\citep{adam_optimizer} with a learning rate of 2e-5 and a batch size of 16. 
We chose proper sequence lengths for each dataset shown in \autoref{tab:datasets}.
It took 1 to 2 days for these models to converge on the validation sets using a V-100 GPU.

\section{Other hyperparameters of FDI}\label{sec:hyperparameters}
For random masking in training, we traversed the document's hierarchy. We randomly masked sentences and then words with 5\% probability.
We then extended each selected word to a non-overlapped $n$-gram with a 50\% probability.
For controllable masking in inference, we set
quantile lower bound $q_{min}$ to 0.4,  
masking sentence's probability $p_s$ to 0.7, 
masking concept's probability $p_c$ to 0.5, 
sentence selection threshold $t_s$ to 0.7, and
max masked rate $\gamma$ to 0.2 for all datasets.

\section{Questionnaire}\label{sec:Questionnaire}
\begin{figure*}[!ht]
  \begin{minipage}{0.5\textwidth}
     \centering
      \includegraphics[width=\columnwidth,keepaspectratio]{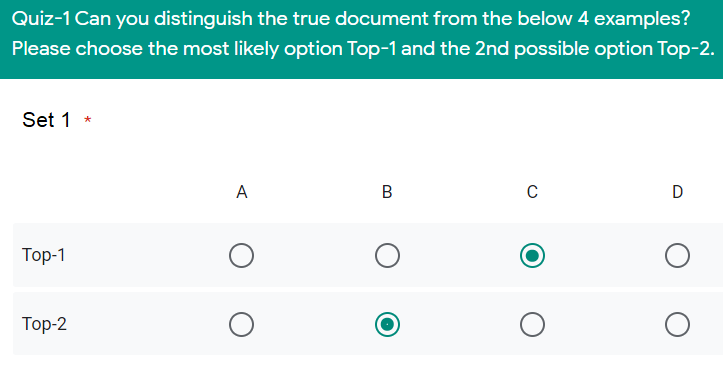}
     \caption{Quiz-1's user interface}
     \label{fig:quiz1 screenshot}
  \end{minipage}\hfill
  \begin{minipage}{0.5\textwidth}
     \centering
      \includegraphics[width=\columnwidth,keepaspectratio]{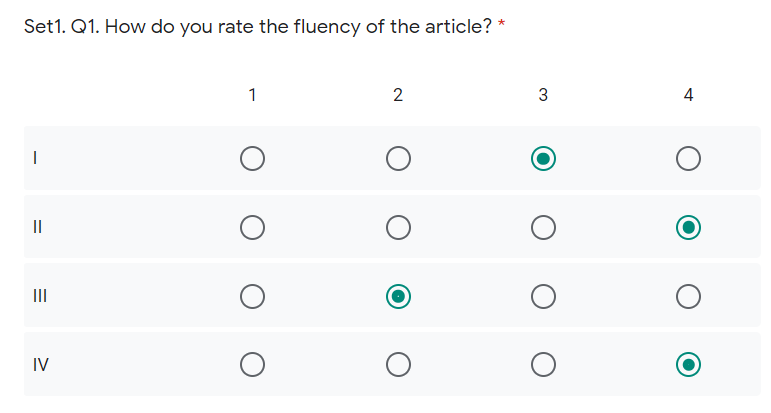}
     \caption{Quiz-2's user interface}
     \label{fig:quiz2 screenshot}
  \end{minipage}
\end{figure*}

\begin{figure*}[!ht]
  \centering
  \includegraphics[width=\linewidth]{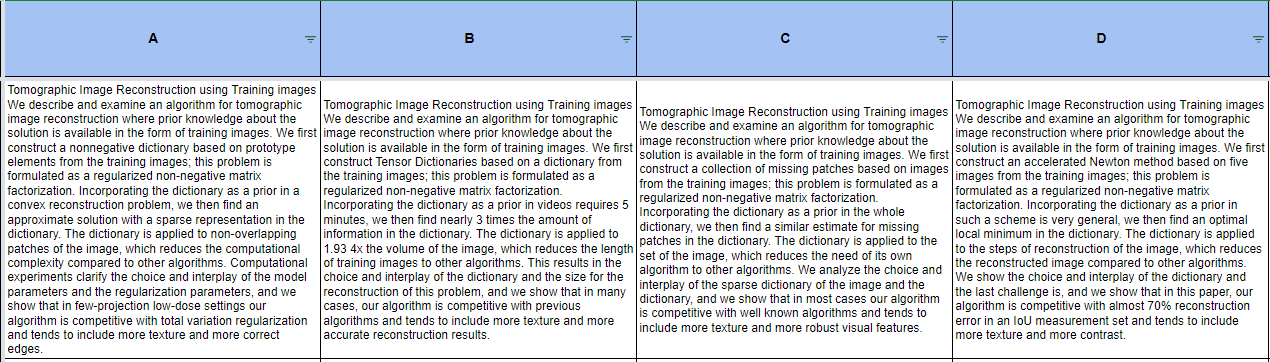} 
  \caption{A Quiz-1's example set consists of 1 true + 3 fake articles generated by an unknown model. 
}
  \label{fig:quiz-1-example}
\end{figure*}

\begin{figure*}[!ht]
  \centering
  \includegraphics[width=\linewidth]{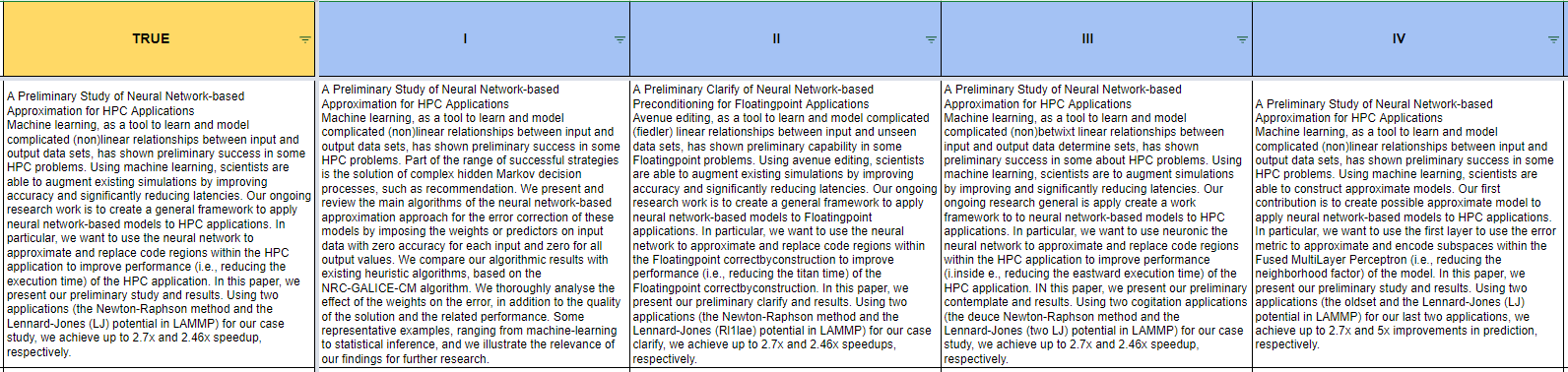}
  \caption{A Quiz-2's example set includes 1 known true document + 4 fake samples generated by 4 models in an unknown order. 
}
  \label{fig:quiz-2-example}
\end{figure*}

\autoref{tab:quiz instructions} explains our quiz's instructions and questions.
\autoref{fig:quiz1 screenshot} and \autoref{fig:quiz2 screenshot} show our designed user interfaces using Google Forms.
\autoref{fig:quiz-1-example} and \autoref{fig:quiz-2-example} shows one example set of Quiz-1 and Quiz-2 in Google Forms, respectively.

\begin{table}[h]
\centering
\small
\begin{tabular}{p{7.5cm}}
\toprule
\textbf{Quiz-1} Assume you are a hacker.
Can you distinguish the true document from the below 4 examples?
Please choose the most likely option Top-1 and the 2nd possible option Top-2.
\\
\\
\textbf{Quiz-2} Assume you are a cyber security expert.
The ideal fake documents should be realistic and provide scalable protective coverage.
They are “close enough” to the original to make the fakes believable, but sufficiently “far enough” to hide and protect private and confidential information. 
Now compared with the true document, would you evaluate the fakeness of the rest four fake samples? 
Below are the questions in details:
\\
\\
Q1. How do you rate the fluency of the article?\\
4. Overall flawless, with only minor typos.\\
3. Non-native, with minor but apparent errors. \\
2. Unnatural/synthetic, the apparent errors affect my reading. \\
1. Incomprehensible, with a lot of corrupted text.                \\
\\
Q2. How do you rate the coherency of the article? Does it make sense?  \\
4. Coherent. There is a logical and consistent relation among the facts presented along the article.\\
3. Partially coherent, I can’t understand what the author means in certain places. \\
2. Somehow confusing, with most parts of the document are confusing. \\
1. I have no (or almost no) idea what the author is trying to say.
\\
\\
Q3. Expert knowledge is required to identify this article is fake.\\
4. Agree, non-expert will find it difficult to distinguish if it is fake.\\
3. Partially agree, expert knowledge may help and speed up this process. \\
2. Somewhat disagree, expert knowledge might not be necessary. \\
1. Disagree, general audience can easily identify it is fake.     
\\
\\
Q4. Is the sample “fake enough”? Does it apply necessary modifications (e.g., insert, replace and delete) to deceive the adversary and protect some essential facts? Note: High scores of fakeness do not mean superiority.\\
4. Excessive. The article may introduce too many changes, substantially diverging from the original topic/fact.
\\
3. Moderate. The article introduces important changes, preserves the coherence, and seems realistic.
\\
2. Marginal. The article introduces changes. However, considerable modifications do not significantly change facts presented in the original document.
\\
1. Inadequate. Only insignificant modifications. \\
\\
Q5. Based on your previous evaluation, how would you rank the fake documents? A good fake copy should look similar to the original document. But what's more important is that it also protects essential information and misleads hackers. Please rank your preference.
(Top-1 the best to Top-4 the worst)
\\
\bottomrule
\caption{The instructions and questions in the Quiz.}
\label{tab:quiz instructions}
\end{tabular}
\end{table}

\end{document}